# Feature Selection Intelligent Algorithm With Mutual Information and Steepest Ascent Strategy

Elkebir Sarhrouni #1, Ahmed Hammouch *, Driss Aboutajdine#
#LRIT, Faculty of Sciences, Mohamed V - Agdal University, Morocco,
1 sarhrouni436@yahoo.fr
* LRGE, ENSET, Mohamed V - Souissi University, Morocco

## Abstract

Remote sensing is a higher technology to produce knowledge for data mining applications. In principle Hyperspectral Images (HSI) is a tool that provides classification remote regions. The HSI contains more than a hundred of images, of the Ground Truth Map (GT). Some images are caring relevant information, but others describe redundant information, or affected bay atmospheric noise. The aim is to reduce dimensionality of HSI. Many studies use mutual information (MI) or MI normalized forms to select appropriate bands. In this paper we design an algorithm based also on MI to and combined with steepest ascent algorithm, to improve a Symmetric Uncertainty coefficient strategy to select relevant bands for classification of HSI. This algorithm is a feature selection tool and a Wrapper strategy. We conduct our study on HSI AVIRIS 92AV3C. This is an artificial intelligent system to control redundancy; we had to clear the difference of the result's algorithm and the human decision, and this can be viewed as case study which human decision is perhaps different to an intelligent algorithm.

## Introduction

For classification substances, the choice of data sets significantly drives the results. But with multivariate data in high dimensionality the problematic will be related at data mining field. To find the good model, we must solve a feature selection problem. This is commonly reencountered when we have $N$ attributes that express $N$ vectors of measures for $C$ classes. The problematic is to find $K$ vectors from the $N$ ones, such as relevant and no redundant; in order to classify substances. The number of selected vectors $K$ must be lower than $N$, the reason is the "Hughes phenomenon" [10]: when $N$ is so large, many cases are needed to detect the relation between the attributes and the classes. No redundant attributes guaranty no complication of learning system and product incorrect prediction [14]. Relevant attributes is there ability to predicate the classes. The Hyperspectral image (HSI), as a set of more than a hundred tow dimensional measurements (images or bands), needs dimensionality reduction. Indeed, the bands don't all contain the information; some bands are irrelevant like those affected by various atmospheric effects, see Figure.3, and decrease the classification accuracy. Finally, there exist redundant bands that must be avoided. We can reduce the dimensionality of hyperspectral images by tow possible scheme: selecting and extraction. Selection picks up o the adequate bands form the existing ones. Extraction generates, from the original bands, a new subset that must contain all information about the classes. This process use some functions, logical or numerical [8][11][9]. Sarhrouni et al [23]. introduced an algorithm based on Mutual Information (MI), reducing dimensionality in two steps: pick up the relevant bands first, and avoiding redundancy second. They illustrate the principia of this algorithm using synthetic bands for HSI AVIRIS 92AV3C [1], see Figure.1 Then they approve its effectiveness with applying it to real data of HIS AVIRIS 92AV3C. Each pixel can be represented as a vector of 220 measurements [7]. Here, we reproduce this approach, and we improve this algorithm bay adding a third step that uses a steepest ascent strategy to pick up the subset of bands adequate.

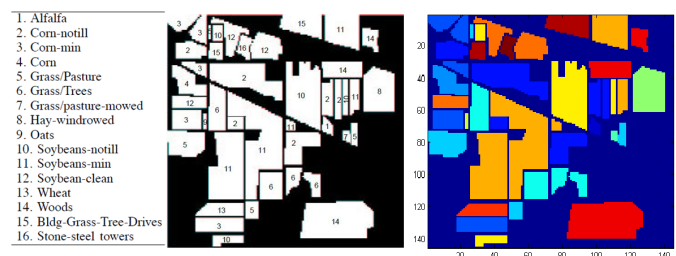

**Figure 1. Ground Truth map of AVIRIS 92AV3C and the 16 classes.**

Hyperspectral image AVIRIS 92AV3C (Airborne Visible Infrared Imaging Spectrometer) [2] is a set of 220 images, called bands, taken over "Indiana Pine" region, at "northwestern Indiana", USA [1]. The 220 measures are related to interval [0.4μm, 2.μ5m]. Each image is a 145 x 145 lines-columns matrix. The ground truth map is given. But only 49% of pixels (10366 pixels) are classified. Each class is designed by a number from 1 to 16. The pixels which the class is unknown are designed bay zeros, Figure.1. The tow dimensional measures are not all well measured, due to the atmospheric effects. Figure.3, this phenomenon on bands 155, 220 and other bands. This HSI presents the problematic of dimensionality reduction

# Mutual Information Based Feature Selection

## A. *Recall of Mutual Information Definition*

Let *A* and *B* are tow ensembles of random variables. The common information between *A* and *B* is expressed as below :

$$I(A,B) = \sum p(A,B) \, log_2 \frac{p(A,B)}{p(A).p(B)}$$

To apply this measure is our study, the ground truth map, and the all bands as closed to random variables. Then we measure their interdependence, according to the demonstration made bay Fano [14]: as soon as mutual information of the selected attributes (or vectors) has high value, the error probability of classification is decreasing as indicated by the formula below:

$$\frac{H(C/X) - 1}{Log_2(N_c)} \leq P_e \leq \frac{H(C/X)}{Log_2}$$

Note that the conditional entropy is related to the mutual information with this formula:

$$H(C/X) = H(C) - I(C;X)$$

Here C is the ground truth map, *X* is the subset of bands candidates. *Nc* is the number of classes. If the features *X* have a higher value of mutual information with the ground truth map, (present more resemblance with the ground truth map), the error probability will be lower. But it is impractical to calculate the conjoint mutual information *I(C,X)* due to the high dimensionality [14].
The Figure .3 shows the MI between the GT and the real bands of HIS AVIRIS 92AV3C [1]
Many studies use a threshold to choose the relevant bands. Guo [3] uses the mutual information to select the top ranking band, and a filter based algorithm to decide if their neighbours are redundant or not. Sarhrouni et al. [17] use also a filter strategy based algorithm on MI to select bands. A wrapper strategy based algorithm on MI, Sarhrouni et al. [18] is also introduced. In Figure .3 we can visually verify this effectiveness of MI to choose relevant features.

## B. Symmetric Uncertainty

This is one of normalized form of Mutual Information; introduced by Witten & Frank 2005 [19]. It's defined as below:

$$U(A,B) = 2.\frac{MI(A,B)}{H(A) + H(B)}$$

*H(X)* is the Entropy of set random variable X. Some studies use this function *U* for recalling images in medical images treatment [9]. Numerous studies use Normalized mutual information [20][21][22]
Figure.3 shows that *Symmetric Uncertainty* means how much information is partaged between *A* and *B* relatively at all information contained in both *A* and *B*:

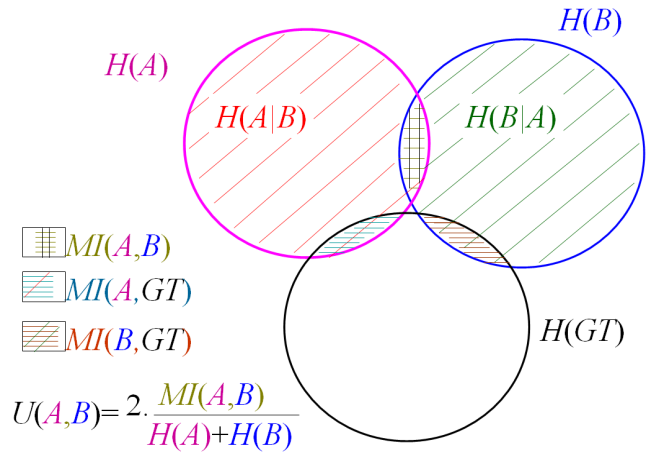

**Figure 2. Illustration of Symmetric Uncertainty.**

## C. Principle of the Method and Algorithm to be improved

Sarhrouni et al [23] have illustrate widely that Symmetric Uncertainly is a measure of redundancy of information between a too bands. We reproduce the algorithm in too steep:
First: Eliminate the no informative bands, because a too noisy bands can have a lowest Symmetric Uncertainty and negatively alter the performance of classification.
Second: Control the redundancy by a threshold of Symmetric Uncertainty between each couple of resulting bands from the first step.
Refer to Sarhrouni et al. [23] for more details.

**Algorithm.1**: From [23]. Band is the HSI. Let Th$_{relevance}$ the threshold for selecting bands more informative,
Th$_{redundancy}$ the threshold for redundancy control.
1) Compute the Mutual Information *(MI)* of the HIS bands and the Ground Truth map GT.
2) Make bands of HIS in ascending order by their *MI* value

3) Cut the bands from HSI that have a lower value than the threshold Th$_{relevance}$, the subset (part of HSI) remaining is *S*.

4) Initialization: $n = length(S)$; $i = 1$, $D$ is a bidirectional array value=1;

//any value greater than 1 can be used: it's useful in step 6)

5) Computation of bidirectional Data $D(n; n)$:

    **for** 1:=1 to n step 1 **do**

        **for** j:=i+1 to n step 1 **do**

        $D(i; j) = U(Band_{S(i)}, Band_{S(j)})$;

// with $U(A;B) = MI(A;B)/(H(A)+H(B))$

        **end for**

    **end for**

//Initialization of the Output of the algorithm

6) $SS = \{\}$ ;

**while** $\min(D) <$ Th$_{redundancy}$ **do**

    // Pick up the argument of the minimum of D

    $(x; y) = argmin(D(.,.))$;

    **if** $\forall L \in SS, D(x; L) <$ Th$_{redundancy}$ **then**

    // x is not redundant with the already selected bands

    $SS = SS \cup \{x\}$

    **end if**

    **if** $\forall L \in SS, D(y; L) <$ Th$_{redundancy}$ **then**

    // y is not redundant with the already selected bands

    $SS = SS \cup \{y\}$

    **end if**

    $D(x; y) = 1; D(x; y) = 1$;

// The cells $D(x; y)$ and $D(y; x)$ will not be checked again

**end while**

7) Finish: The final subset SS contains bands from HSI according to the couple of thresholds:

    (Th$_{hrelevance}$, Th$_{redundancy}$).

## C. Results of the Method and Algorithm to be improved

We apply the proposed algorithm on the hyperspectral image AVIRIS 92AV3C [1], 50% of the labelled pixels are randomly chosen and used in training; and the other 50% are used for testing classification [3]. The classifier used is the SVM [5][12][4].
Algorithm 1 shows more details.

From HSI, we can eliminate the no informative ones, bay a thresholding, see the proposed algorithm. Figure.3 gives the MI of the HSI AVIRIS 92AV3C with the ground truth map (GT)

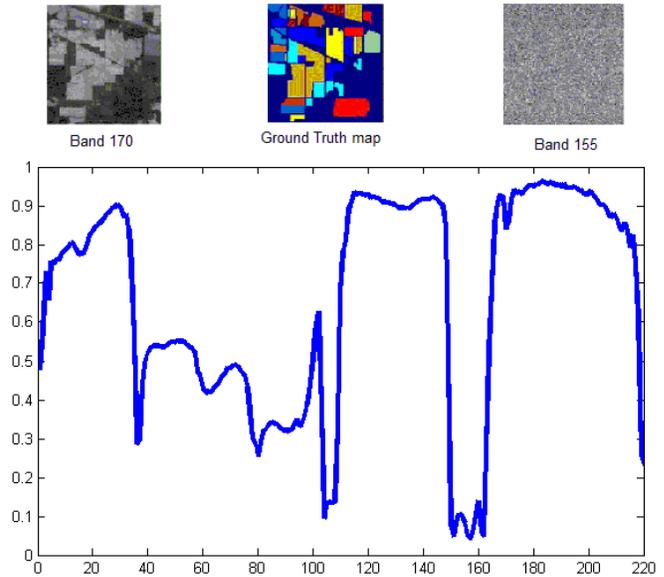

**Fig. 3. Mutual information of AVIRIS with the Ground Truth map.**

From the remaining subset bands, we must eliminate redundant ones using the proposed algorithm. Table I gives the accuracy off classification for a number of bands with several thresholds.

HUMAN DECISION:

An observer, however intelligent, can fragment the results in Table. I in six zones of couple values of thresholds (*TH,IM*):

Zone1: This is practically no control of relevance and no control of redundancy. So, there is no action of the algorithm.

Zone2: This is a hard selection: a few more relevant and no redundant bands are selected.

Zone3: This is an interesting zone. We can easily have 80% of classification accuracy with about 40 bands.

Zone4: This is the very important zone; we have the very useful behaviours of the algorithm. For example, with a few numbers of bands 19 we have classification accuracy 80%.

Zone5: here we make a hard control of redundancy, but the bands candidates are nearer to the GT, and they may be more redundant. So, we can't have interesting results.

Zone6: when we do not control properly the relevance, some bands affected bay transfer affects may be non redundant, and can be selected, so the accuracy of classification is decreasing.

This algorithm is very effectiveness for redundancy and relevance control, in feature selection area.

The question now is how to find a road to the suboptimal (or optimal) solution, automatically, without computing this entire table? This will be the proposed improvement of this paper.

| | MI: Threshold for control the relevence (MI of bands with Ground Truth) | | | | | | | | | | | | | | | |
|---|---|---|---|---|---|---|---|---|---|---|---|---|---|---|---|---|
| | MI = 0 | | MI > 0,4 | | MI > 0,45 | | MI > 0,57 | | MI > 0,6 | | MI > 0,9 | | MI > 0,91 | | MI > 0,93 | |
| TH | N.B | ac(%) | N.B | ac(%) | N.B | ac(%) | N.B | ac(%) | N.B | ac(%) | N.B | ac(%) | N.B | ac(%) | N.B | ac(%) |
| 0,10 | 26 | 37,68 | 3 | 46,31 | 3 | 46,62 | - | - | - | - | - | - | - | - | - | - |
| 0,15 | 30 | 44,40 | 6 | 49,52 | 6 | 60,65 | 2 | 43,74 | 2 | 43,25 | - | - | - | - | - | - |
| 0,20 | 34 | 45,31 | 10 | 64,70 | 10 | 65,44 | 6 | 59,69 | 5 | 51,61 | - | - | - | - | - | - |
| 0,25 | 40 | 46,44 | 13 | 67,27 | 13 | 68,13 | 8 | 56,63 | 8 | 61,82 | - | - | - | - | - | - |
| 0,30 | 47 | 47,50 | 20 | 75,32 | 18 | 74,67 | 15 | 66,63 | 14 | 67,60 | - | - | - | - | - | - |
| 0,35 | 59 | 47,13 | 29 | 77,77 | 29 | 77,67 | 25 | 73,33 | 23 | 71,03 | 3 | 65,58 | 2 | 55,48 | - | - |
| 0,40 | 70 | 46,76 | 40 | 81,41 | 38 | 80,34 | 32 | 77,48 | 30 | 77,34 | 5 | 71,73 | 4 | 63,80 | - | - |
| 0,43 | 78 | 46,35 | 47 | 83,21 | 46 | 82,82 | 37 | 77,21 | 32 | 77,40 | 6 | 73,29 | 4 | 63,80 | - | - |
| 0,45 | 90 | 45,92 | 56 | 84,08 | 54 | 83,44 | 45 | 80,65 | 41 | 80,19 | 11 | 75,86 | 7 | 65,69 | 2 | 52,09 |
| 0,46 | 93 | 45,68 | 61 | 84,69 | 58 | 84,32 | 48 | 81,45 | 44 | 80,89 | 13 | 78,28 | 10 | 69,02 | 2 | 52,09 |
| 0,47 | 102 | 45,12 | 66 | 85,25 | 61 | 84,57 | 52 | 82,10 | 50 | 81,38 | 16 | 79,82 | 12 | 71,75 | 2 | 52,09 |
| 0,48 | 109 | 45,10 | 70 | 86,42 | 65 | 85,43 | 56 | 82,04 | 54 | 82,14 | 19 | 80,38 | 14 | 72,73 | 3 | 55,19 |
| 0,49 | 115 | 44,77 | 76 | 86,89 | 71 | 86,19 | 61 | 83,07 | 59 | 82,74 | 23 | 81,39 | 18 | 74,38 | 3 | 55,19 |
| 0,50 | 121 | 44,57 | 80 | 86,69 | 75 | 86,40 | 64 | 83,97 | 62 | 83,73 | 25 | 81,73 | 20 | 75,04 | 4 | 56,65 |
| 0,51 | 127 | 44,15 | 87 | 87,51 | 82 | 87,14 | 68 | 84,34 | 65 | 84,40 | 29 | 82,70 | 23 | 75,67 | 6 | 57,82 |
| 0,52 | 135 | 43,74 | 89 | 87,59 | 84 | 87,38 | 73 | 85,02 | 71 | 85,04 | 30 | 83,28 | 25 | 76,89 | 7 | 58,08 |
| 0,53 | 141 | 43,46 | 96 | 87,63 | 91 | 87,43 | 75 | 85,21 | 74 | 84,86 | 32 | 82,93 | 27 | 77,36 | 10 | 59,75 |
| 0,54 | 147 | 43,23 | 104 | 87,94 | 99 | 87,61 | 81 | 85,82 | 79 | 85,41 | 39 | 83,99 | 34 | 78,32 | 13 | 60,78 |
| 0,55 | 154 | 42,55 | 108 | 87,94 | 103 | 87,78 | 84 | 86,23 | 82 | 85,68 | 40 | 83,64 | 35 | 78,67 | 15 | 61,84 |
| 0,56 | 158 | 42,35 | 110 | 87,78 | 105 | 87,63 | 85 | 86,07 | 83 | 86,11 | 42 | 84,16 | 36 | 78,49 | 16 | 61,72 |
| 0,70 | 220 | 38,96 | 173 | 88,72 | 163 | 88,41 | 128 | 87,88 | 126 | 87,55 | 67 | 86,71 | 54 | 81,77 | 22 | 63,72 |
| 0,90 | 220 | 38,96 | 173 | 88,72 | 163 | 88,41 | 128 | 87,88 | 126 | 87,55 | 67 | 86,71 | 54 | 81,77 | 22 | 63,72 |
| 1,00 | 220 | 38,96 | 173 | 88,72 | 163 | 88,41 | 128 | 87,88 | 126 | 87,55 | 67 | 86,71 | 54 | 81,77 | 22 | 63,72 |

TH: Threshold for control the redundancy

N,B : Number of Banbds retained for the couple of threshold (MI,TH)
ac(%) :The accuracy of classification calculated for the couple of threshold (MI,TH)

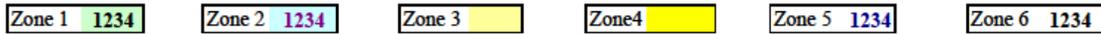

Zone 1  1234    Zone 2  1234    Zone 3    Zone4    Zone 5  1234    Zone 6  1234

TABLE.I The Symmetric Uncertainty of the relevant synthetic bands

# The Improved Algorithm Proposed

Here we propose to apply the steepest ascent search strategy, over the space formed bay the bands resulting from the first step of Algorithm .1. To find the adequate subset of bands automatically.

## A. Principle of Steepest Ascent Algorithm Applied to HSI.

We propose to search for suboptimal solutions that are constrained local maxima of the classification accuracy. So we introduce here a form of Steepest Ascent adapted for HSI, and principally we define a criterion with operators according to the association rules, as below:

**Definition.1:** We define a formalism of criterion, for moving from a point (TH—MI) to others as below:

**Rule.1:** If the moving makes accuracy decreasing and increases the number of bands, it will not be done. We note this operator: **J-not**

**Rule.2:** If the several directions increase the accuracy and decrease the umber of bands needed (or let them constant), we retain the direction with the highest ratio R= Absolute Variation of Accuracy by Band. We note this operator: **J-best.**

**Rule.3:** If the moving decreases the accuracy (or lets it constant) and decreases number of bands, we retain the direction with the lowest ratio R= Absolute Variation of Accuracy by Band. We note this operator **J-lost.** (With accuracy constant and increasing bands number we sill have J-lost and R=0).

**Rule.4:** If the moving increases accuracy and increases number of bands (or lets it constant), we retain the direction with the highest ratio R= Absolute Variation of Accuracy by Band. We note this operator: **J-great.**

**Rule.5: J-best** overrides **J-great**, **J-great** overrides **J-lost**.

**Rule.6:** If a point is not defined, it corresponds to **J-not**.

**Rule.7:** The retreat is interdicting (**Int**).

**Definition.2:** We say that a pair (TH—MI) is a local maximum if the operator **J-not** appears for all its neighbours, or another operator points to an already visited pair (TH--MI). In this case the bands subset related to the threshold will be retained as suboptimal solution.

The Algorithm.2 implements the proposed SA based method.

**Algorithm.2**: The conditions of **Algorithm.1** are verified. Also, we use **Algorithm.1** as a procedure.

1)**First-Action:** We start from a point corresponding to an initial subset of bands of HIS, i.e. we start from a pair of Thresholds ($Th_{hrelevance}$, $Th_{redundancy}$), we extract the bands subset S0 using **Algorithm.1**, and we compute Classification accuracy C_S0 .
2)**Second Action:** We repeat First-Action for all Neighboring threshold couples of ($Th_{hrelevance}$, $Th_{redundancy}$).
3)**Third Action:** We move to the point (threshold couple) according to the operator criterion mentioned above.
4)**Forth-Action: Repeat** the **Second-Action** for the new threshold couple **until** the operator J-not appear for all neighbours of the couple ($Th_{hrelevance}$, $Th_{redundancy}$).
5)**end of Algorithm.2**

To better explore the space of solutions, this algorithm must be run several times with randomized initializations, because another local maximum may be obtained at each run

This strategy differs from the other search algorithms for feature selection, which usually progressively increase (e.g., SFS) or decrease (e.g., SBS) the number of bands in, with possible backtracking (e.g., SFFS and SFBS).

## B. Results and Comments

Table. II shows three trajectories of the evolution over the couples selected, according to three randomized initializations of the algorithm. Tow of them point to the same local maximum.

.

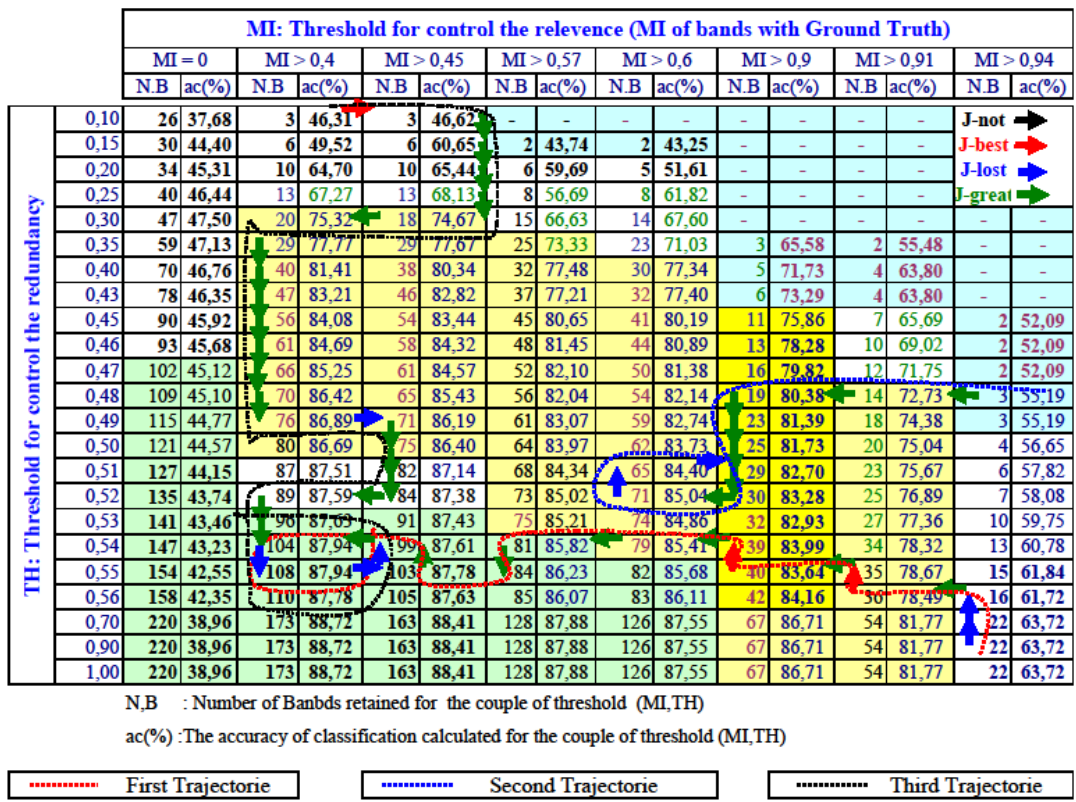

TABLE.III   Trajectories of the Evolution Over the Couples Selected According to Three randomized Initializations

Table. III gives the transitions of the first trajectory, for the first random start point.

|  | Point TH--MI | Direction | | | |
|---|---|---|---|---|---|
|  |  | Left | Down | Top | Ritgh |
|  |  | R= Absolute Variation of Accuracy by Band | | | |
| ↓ | 0,90--0,94 | J-lost- R=0,56 | J-lost R=0,00 | J-lost-R=0,00 | J-not |
| ↓ | 0,70--0,91 | J-lost-R=0,56 | Int | J-lost-R=0,33 | J-not |
| ↓ | 0,56--0,94 | J-gret-R=0,83 | Int | J-lost-R=0,12 | J-not |
| ↓ | 0,56--0,91 | J-gret-R=0,18 | J-gret-R=0,94 | J-best-R=0,18 | Int |
| ↓ | 0,55--0,91 | J-gret-R=0,99 | Int | J-lost-R=0,35 | J-lost-R=0,84 |
| ↓ | 0,55--0,90 | J-gret-R=0,04 | J-gret-R=0,26 | J-best-R=0,35 | Int |
| ↓ | 0,54--0,90 | J-gret-R=0,03 | Int | J-lost-R=0,15 | J-lost-R=1,13 |
| ↓ | 0,54--0,60 | J-gret-R=0,20 | J-gret-R=0,09 | J-lost-R=0,11 | Int |
| ↓ | 0,54--0,57 | J-gret-R=0,09 | J-gret-R=0,41 | J-lost-R=0,10 | Int |
| ↓ | 0,55--0,57 | J-gret-R=0,02 | J-not | Int | J-lost-R=0,27 |
| ↓ | 0,55--0,45 | J-gret-R=0,03 | J-not | J-lost-R=0,04 | Int |
| ↓ | 0,54--0,45 | J-gret-R=0,06 | Int | J-lost-R=0,02 | J-lost-R=0,09 |
| ↑ | 0,54--0,40 | J-not | J-lost R=0,00 | J-lost-R=0,03 | Int |
| ↓ | 0,55--0,40 | J-not | J-not | Int | J-lost R=0,03 |
| ↓ | 0,55--0,45 | Int | J-not | J-lost-R=0,04 | J-lost R=0,17 |
|  | 0,54--0,45 | J-great R=0,06 | Int | J-lost-R=0,02 | J-not |

J-not → (black)
J-best → (red)
J-lost → (blue)
J-great → (green)
Int    Interdiction of retreat is equivalent to J-not

Table of Transitions for the First Trajectory with the Operators

TABLE.IIII    Transitions of the First Trajectory

Table. IV gives the transitions of the second trajectory, for the second random start point:

|  | Point TH--MI | Direction | | | |
|---|---|---|---|---|---|
|  |  | Left | Down | Top | Ritgh |
|  |  | R= Absolute Variation of Accuracy by Band | | | |
| ↓ | 0,48--0,96 | J-not | J-gret R=1,36 | J-lost-R=3,1 | J-not |
| ↓ | 0,48--0,91 | J-gret R=1,53 | J-gret R=0,74 | J-lost-R=0,49 | Int |
| ↓ | 0,48--0,90 | J-gret-R=0,07 | J-gret R=0,25 | J-lost-R=0,18 | Int |
|  | 0,49--0,90 | J-gret-R=0,03 | J-great-R=0,17 | Int | J-lost-R=0,18 |
|  | 0,50--0,90 | J-gret-R=0,05 | J-gret-R=0,24 | Int | J-lost-R=1,33 |
| ↑ | 0,51--0,90 | J-gret-R=0,04 | J-gret-R=0,58 | Int | J-lost-R=1,17 |
| ↓ | 0,52--0,90 | J-gret-R=0,04 | J-not | Int | J-lost-R=1,12 |
|  | 0,52--0,60 | J-not | J-not | J-lost-R=0,22 | Int |
|  | 0,51--0,60 | J-not | Int | J-lost-R=0,22 | J-lost-R=0,04 |

J-not → (black)
J-best → (red)
J-lost → (blue)
J-great → (green)
Int    Interdiction of retreat

Table of Transitions for the Second Trajectory with the Operators

TABLE.IVI    Transitions of the Second Trajectory

And finally, the Table. V gives the transitions for the tired trajectory, for the third random point.

| Point TH--MI | Direction | | | |
|---|---|---|---|---|
| | Left | Down | Top | Ritgh |
| | R= Absolute Variation of Accuracy by Band | | | |
| 0,10--0,40 | J-not | J-great R=1,07 | J-not | J-best |
| 0,10--0,45 | Int | J-great R=4,67 | J-not | J-not |
| 0,15--0,45 | J-lost R=+∞ | J-great R=1,19 | Int | J-not |
| 0,20--0,45 | J-lost R=+∞ | J-great R=0,89 | Int | J-lost R=1,43 |
| 0,25--0,45 | J-lost R=+∞ | J-great R=1,30 | Int | J-lost R=2,28 |
| 0,30--0,45 | J-great R=0,32 | J-great R=0,27 | Int | J-lost R=2,68 |
| 0,30--0,40 | J-not | J-great R=0,27 | J-lost R=1,15 | J-not |
| 0,35--0,40 | J-not | J-great R=2,45 | Int | J-lost R=+∞ |
| 0,40--0,40 | J-not | J-great R=0,16 | Int | J-lost R=0,53 |
| 0,43--0,40 | J-not | J-great R=0,09 | Int | J-lost R=0,39 |
| 0,45--0,40 | J-not | J-great R=0,12 | Int | J-lost R=0,32 |
| 0,46--0,40 | J-not | J-great R=0,11 | Int | J-lost R=0,12 |
| 0,47--0,40 | J-not | J-great R=0,29 | Int | J-lost R=0,13 |
| 0,48--0,40 | J-not | J-great R=0,07 | Int | J-lost R=0,19 |
| 0,49--0,40 | J-not | J-not | Int | J-lost R=0,14 |
| 0,49--0,45 | Int | J-great R=0,05 | J-lostt R=0,12 | J-lost R=0,31 |
| 0,50--0,45 | J-great R=0,05 | J-great R=0,10 | Int | J-lost R=0,22 |
| 0,51--0,45 | J-great R=0,07 | J-great R=0,12 | Int | J-lost R=0,20 |
| 0,52--0,45 | J-great R=0,04 | J-great R=0,007 | Int | J-lost R=0,26 |
| 0,52--0,40 | J-not | J-great R=0,005 | J-lost R=0,04 | Int |
| 0,53--0,40 | J-not | J-great R=0,03 | Int | J-lost R=0,04 |
| 0,54--0,40 | J-not | J-lost-R=0,00 | Int | J-lost R=0,06 |
| 0,55--0,40 | J-not | J-not | Int | J-lost R=0,03 |
| 0,55--0,45 | Int | J-not | J-lost-R=0,04 | J-lost R=0,17 |
| 0,54--0,45 | J-great R=0,06 | Int | J-lost-R=0,02 | J-not |

Table of Transitions for the Third Trajectory with the Operators

TABLE.VI   Transitions of the Third Trajectory

We can note the effectiveness performance of this algorithm, but a human observatory can prefer for example a choice like (0.56--0.9): so the algorithm.1 gives the 42 bands, and with the classifier SVM [5][12] [4], 50% of the labeled pixels are randomly chosen and used in training; and the other 50% are used for testing classification [3], the accuracy of classification is 84.16%.

This is a paradigm problem. This is a subjective choice. In the proposed method the rules are the artificial intelligent part, and they ply a major role; principally the measure of the importance of direction by the ratio R, is deterministic for the algorithm performance.

# Conclusion

In the data mining field, the features selection in high dimensionality, like HIS, plies a deterministic role. The problematic is always open. Some methods and algorithms have to select relevant and no redundant subset features. In this paper improved one algorithm that processes separately the relevance and the redundancy, to automatically choose a suboptimal solution. It's based on the steepest ascent method, and mutual information. We apply our method to classify the region Indiana Pin with the Hyperspectral Image AVIRIS 92AV3C. This algorithm is a Wrapper strategy (i.e., it calls the classifier during the selection). In the first times we use mutual information to pick up relevant bands by thresholding (like most method already used). In the second times, we perform a search strategy based on steepest ascent method, symmetric uncertainty, and mutual information. We conclude the effectiveness of our algorithm to select the relevant and no redundant bands. However, there are no guaranties that the chosen bands are the optimal ones, because some redundancy can be important to reinforcement of learning classification system. In general, this is an important process that can be more improved to be less expensive.